# Improving Gradient Estimation by Incorporating Sensor Data

**Gregory Lawrence**
Computer Science Division
U.C. Berkeley
gregl@cs.berkeley.edu

**Stuart Russell**
Computer Science Division
U.C. Berkeley
russell@cs.berkeley.edu

## Abstract

An efficient policy search algorithm should estimate the local gradient of the objective function, with respect to the policy parameters, from as few trials as possible. Whereas most policy search methods estimate this gradient by observing the *rewards* obtained during policy trials, we show, both theoretically and empirically, that taking into account the *sensor data* as well gives better gradient estimates and hence faster learning. The reason is that rewards obtained during policy execution vary from trial to trial due to noise in the environment; sensor data, which correlates with the noise, can be used to partially correct for this variation, resulting in an estimator with lower variance.

## 1 INTRODUCTION

Policy search algorithms have been very effective in learning good policies in the reinforcement learning setting. Successful applications include helicopter flight [8], quadruped locomotion [5], and baseball hitting [10]. These methods work by adjusting the parameters of a policy to improve its *value*, i.e., the expected sum of rewards (possibly discounted) obtained during policy execution. To do this, the algorithms repeatedly estimate the *gradient* of the value with respect to the parameters, using information observed during policy trials, and then adjust the parameters in the "uphill" direction. Because trials can be expensive, especially in physical environments, a number of authors have presented techniques to reduce the number of required trials—mainly by reducing the variance of the gradient estimator [1, 3, 6, 7, 10, 9, 11, 13].

Generally speaking, these methods estimate the gradient from the policy parameter settings on each trial and the *score* (the actual sum of rewards), ignoring the sensor data.[1] The main point of this paper is that the sensor data obtained during each trial also provides a useful signal that can reduce the variance of gradient estimators. To understand how this may be so, consider first a case in which it is *not* so: that is, the noise-free case where the score is a deterministic function of the policy parameters. In that case, the local gradient can be estimated exactly from a small set of trials with policy parameter settings closely spaced around the setting of interest, and the sensor data can provide no more information.[2] In the noisy case, however, a gradient estimator can be easily misled by trials of bad policies that yield fortuitously good scores and *vice versa*. In essence, what we propose is that sensor data can account, at least partially, for the *deviation* in the score of each trial from its expected value. Conditioned on the sensor data, therefore, the posterior estimate for the policy value will be closer to the true value.

As an example, consider the problem of firing a cannon at a distant target (Figure 1). Imagine that, after firing several shots that fall short, you increase the firing angle to $\theta = \theta_0$. The next shot sails over the target. Normally, you would decrease $\theta$ again. Suppose, however, that the sound of this last shot was much louder than usual; this suggests that the muzzle velocity was higher than intended and accounts for the poor outcome. It might therefore be sensible to stay at $\theta_0$ for the next shot. Notice, though, that this decision re-quires some knowledge of the relationship between the sensor data (loudness of bang) and the score; learning such relationships is a key element of our method.

[1] Of course, the policies themselves may use the sensor data to select actions. One might also interpret the approach of [6] as using observed perturbations to improve gradient estimation—but only for the restricted case of perfect sensing of noise applied directly to the policy parameters. The current paper removes these unrealistic assumptions, allowing the method to apply to any real physical system; it also provides a more general explanation of the benefits of sensor data for gradient estimation.

[2] If we are willing to step outside the policy search framework, of course, then we can use the sensor data to learn a transition model from which the optimal policy can be obtained by offline methods.

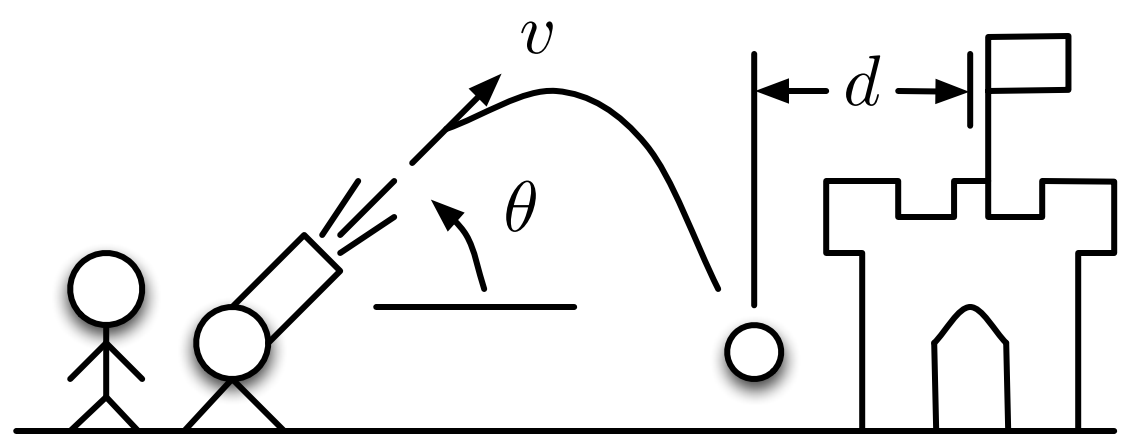

Figure 1: The cannon problem.

The remainder of the paper is organized as follows. Section 2 introduces the basic technical approach: instead of estimating the gradient by fitting a local linear model for the value as a function of the policy parameters, we fit a linear model of the value as a function of the policy parameters and the (transformed) sensor values. We show that estimator variance is minimized if the sensor data is transformed by a projection that renders it statistically independent of the policy parameters while remaining correlated with the perturbations in the score; we also present a practical technique that approximates this ideal transformation. Section 3 describes how we applied these techniques to a dart throwing task and a quadruped locomotion task, in both cases obtaining improved learning curves compared to a method that does not use the sensor data. Section 4 discusses possibilities for future improvements.

## 2 INCORPORATING SENSOR INFORMATION

A policy $\pi$ determines how an agent chooses its actions given its past observations, and the reinforcement learning goal is to find a policy $\pi^*$ that maximizes the performance. Each policy trial generates a *history* $h$, a sequence of percept-action-reward tuples; the *score* $f(h)$ is the sum of the reward values. The optimal policy $\pi^*$ maximizes the value function $V(\pi) = \mathrm{E}[f(\mathbf{h})|\pi]$ where the histories $\mathbf{h}$ are generated from $\pi$. In this paper we perform policy search by hill-climbing through a space of parameterized policies $\pi \in \mathbb{R}^d$. We climb through this space by adjusting our current policy $\pi_0$ in the direction of the gradient $\nabla_\pi V(\pi)|_{\pi=\pi_0}$. At each hill-climbing step, the gradient is estimated from $n$ policy trials where we vary the parameters of each trial for exploration purposes.

### 2.1 TOY EXAMPLE

To illustrate the main contributions of this paper, we will examine the toy cannon problem (Figure 1). Here, the policy $\pi = (\pi_v, \pi_\theta)^T$ consists of a desired cannon angle, $\pi_\theta \in [0, \pi/2]$ and a desired initial velocity $\pi_v > 0$. Following [6], we assume the policy itself is perturbed by noise to give the actual controls $u = (u_v, u_\theta)^T$. We assume that the agent has access to a noisy sensor that measures the perturbation $(u - \pi)$ and let $s = (s_v, s_\theta)$ denote its value. There is additive, zero-mean Gaussian noise in both the control and sensor values; the control noise has covariance matrix $\Sigma_u$ and the sensor noise has covariance matrix $\Sigma_s$. The history $h$ for this problem has a single tuple with the desired action $\pi$, the sensor value $s$, and the reward $-d^2$, where $d$ is the distance from the target to where the cannon ball lands. The score is just the reward in this case, so maximizing $V(\pi) = \mathrm{E}[f(\mathbf{h})|\pi]$ is equivalent to minimizing the expected squared distance error.

To demonstrate the benefits of incorporating the sensor data, suppose that we had a perfect sensor. In that case, we have $u = \pi + s$. Furthermore let us assume that $f(h)$ is (locally) linear in the actual control $u$. Then the score function can be written as $f(\pi, s) = (\pi + s)^T A_{\pi_0} + b_{\pi_0}$. A typical approach to gradient estimation in this setting would be to ignore the sensor data and use samples of $\pi$ and $f(\pi, s)$ to fit a local linear approximation to $V$ around $\pi_0$:

$$\hat{V}(\pi) = \pi^T A_{\pi_0} + b_{\pi_0} \ .$$

Using least-squares regression, we estimate the gradient by learning $A_{\pi_0}$ from $n$ policy trials. The control noise causes perturbations in the scores with variance given by ${A_{\pi_0}}^T \Sigma_u A_{\pi_0}$. Hence, the estimator $A_{\pi_0}$ will not be exact and we may need many samples to be confident in our estimate of the gradient.

To obtain the benefits of incorporating the sensor data, we can instead fit a sensor-data-dependent linear approximation to the scoring function $f$ itself (rather than its expectation $V$):

$$\hat{f}(\pi, s) = \pi^T A_{\pi_0} + s^T A_s + b_{\pi_0} \ .$$

From samples of $\pi$, $f(\pi, s)$, and $s$ we can use linear regression to learn both $A_{\pi_0}$ and $A_s$ in this equation. The value that we learn for $A_{\pi_0}$ can be used as our estimate of the gradient of V. In the case of perfect sensing, $\hat{f}(\pi, s)$ can be a locally *exact* fit because the score will be a deterministic function of $\pi$ and $s$. Thus, we have a perfect gradient estimator. Intuitively, the more informative the sensor, the better our gradient estimator.

There are two important issues to note with this analysis. As we will see in Section 2.2, the estimator $A_{\pi_0}$

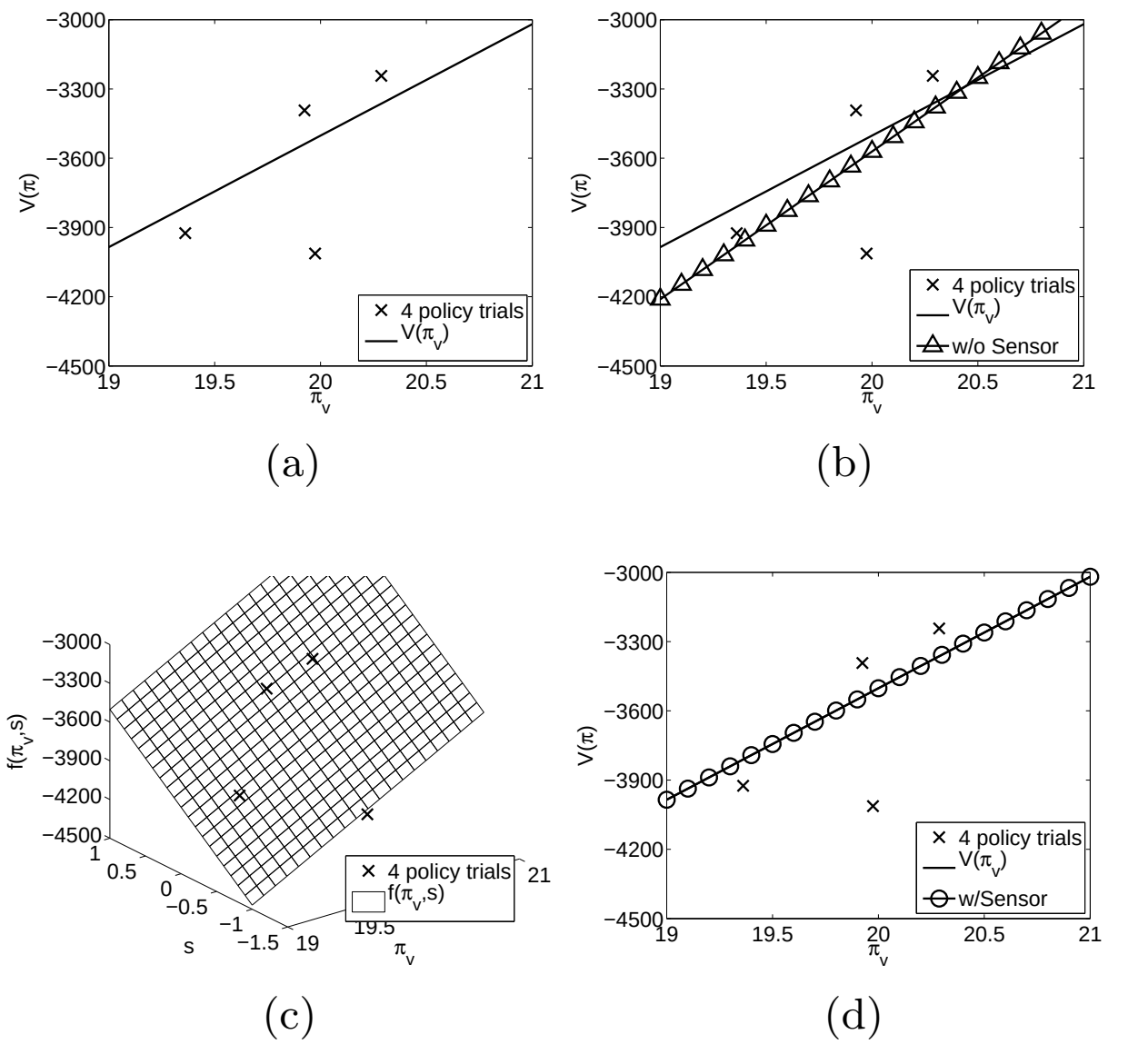

Figure 2: (a) The true value function with four policy trials superimposed. (b) Least squares fit of these four trials. (c) Least squares fit of the the four trials as a function of $\pi$ and $s$. (d) Projection of the fitted plane from (c) to give a linear estimator for $V(\pi)$ (circles). Note that the estimator fits the true value function exactly.

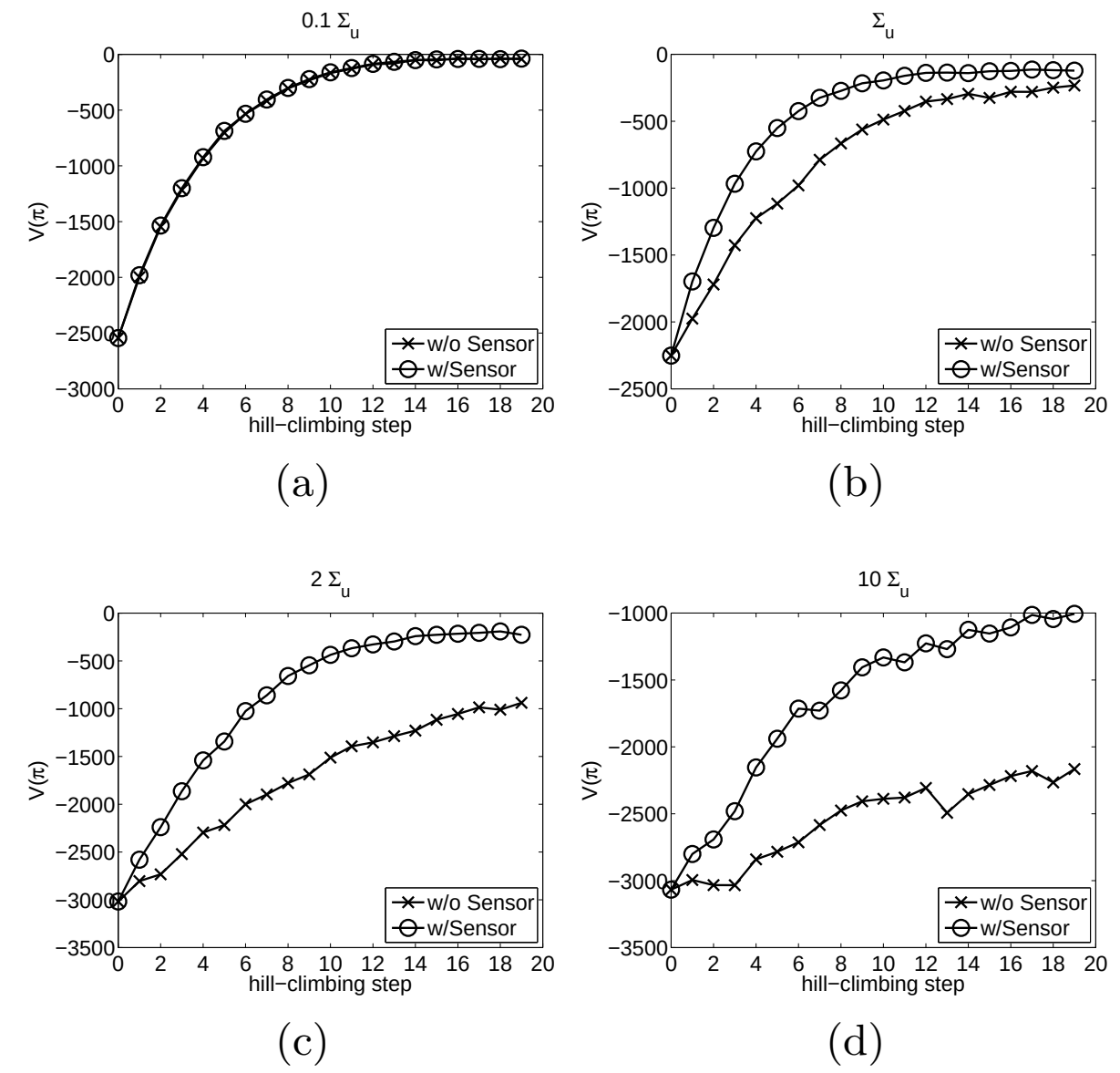

Figure 3: The learning curve performances of two difference policy search algorithms as we increase the level of actuator noise. One curve estimates the gradient while ignoring the sensor data and the other curve estimates the gradient using the sensor data. As mentioned in Section 1, using the sensor data becomes more beneficial as the level of noise in the score increases.

will be unbiased *provided* the sensor data are independent of the policy parameters. Therefore, it will be advantageous to project the sensor data in such a way as to render it independent of the policy parameters to the extent possible, while maintaining its correlation with the perturbations in the score. Second, the sensor-data-dependent fit for $\hat{f}$ requires extra parameters to be learned for $A_s$ (two extra parameters in our example), which may in turn require extra trials. We will see in the next subsection that the relative efficiency of this estimator depends on the amount of noise in the observed scores that can be corrected for using the sensor data and the number of extra parameters to be learned.

Figures 2(a-d) demonstrates how incorporating the sensor data is beneficial. For ease of graphical depiction, we consider a restricted version of the cannon problem in which $u_\theta$ is fixed at 45 degrees. We consider a linear approximation to the score around the nominal cannon velocity $\pi_{v0} = 20\text{ms}^{-1}$. Figure 2(a) shows the true value function with four policy trials superimposed. Figure 2(b) shows the least squares fit of these four trials. The slope of this line is an estimate of the gradient of $V(\pi)$. Figure 2(c) shows the least squares fit of the four trials as a function of $\pi$ and $s$ and Figure 2(d) shows this least squares fit projected onto the plane that spans the $\pi_v$ and $f(h)$ axes. Notice that it perfectly fits the true value. This is because deviations from the expected score are explained away by the sensor values. We expand upon this in the next section.

Figure 3 shows different learning curves for the cannon problem as we increase the level of actuator noise ($\Sigma_u = \left[\begin{smallmatrix} 1 & 0 \\ 0 & 4 \end{smallmatrix}\right]$). At each hill-climbing step we ran 10 trials, each with a different policy in the neighborhood of the current policy, and we averaged the resulting learning curve over 100 hill-climbing runs. One curve shows the learning performance achieved while ignoring the sensor data while the other curve shows what happens when incorporating the sensor data. Notice that the performance gains from using sensor data become more pronounced as the noise level increases.

## 2.2 VARIANCE REDUCTION

In this section we compare the variance of a gradient estimator that incorporates sensor data to an estimator that ignores it. We assume that the score function is an unknown linear function of both the policy parameters and the sensor data. The gradient estimators take $n$ policy trials from the current hill-climbing step and return an estimate of $\nabla_\pi V(\pi)|_{\pi=\pi_0}$.

We will show that an agent can reduce the variance of its gradient estimates by choosing a sensor encoding that correlates with the noise-induced perturbations in the score. To get an unbiased estimate, the sensor data must be uncorrelated with $\pi$. In Section 2.2.3 we give expressions for the bias and variance of a gradient estimator in the correlated setting. A biased estimator that incorporates sensor data may still outperform one that ignores it as long as the bias remains small.

Let $\Pi = [\pi_1, \ldots, \pi_n]^T$ be the matrix of policy parameters and let $S = [s_1, \ldots, s_n]^T$ be a matrix whose rows contain the corresponding sensor values. We assume that the sensors have a Gaussian distribution with variance $\Sigma_s$ and mean $\mu_s$. Let $f = [f_1, \ldots, f_n]^T$ be a column vector whose entries are the score values and let $w = [w_1, \ldots, w_n]$ be a column vector of zero-mean noise which is added to the output with variance ${\sigma_w}^2$. The score is written as $f_2(\pi, s, w) = \pi^T A_\pi + s^T A_s + b + w$. The scores for the current set of policy trials are given by the following equation:

$$f = \Pi A_\pi + S A_s + \mathbf{1}_n b + w, \quad (1)$$

where $\mathbf{1}_n$ is a column vector of ones.

Each estimator learns a local linear model of the scoring function using linear regression on the data obtained from $n$ policy trials. For exploration purposes the policy parameters of each trial are assumed to be distributed around a nominal policy $\pi_0$ according to a Gaussian distribution with variance $\Sigma_e$. We assume that the policy parameters, scores, and sensor data have been centered around zero.

### 2.2.1 Ignore Sensor Data

From the point of view of a gradient estimator that ignores the sensor data, additional noise will appear to be added to the scores that will not be explained by perturbations in the sensor data. Let $v$ represent this noise where each element is given by the equation $v = s^T A_s + w$. The variance of each entry in $v$ is given by the expression ${A_s}^T \Sigma_s A_s + \sigma^2$. The score function $f_1(\pi, v) = \pi^T A_\pi + b + v$ is equivalent in value to $f_2$. We can learn the linear relationship between the policy parameters and the score by performing linear regression on the set of $n$ policy trials. In other words we find a suitable estimate for $A_\pi$ in the following equation:

$$f = \Pi A_\pi + \mathbf{1}_n b + v. \quad (2)$$

We are interested in the gradient of $\mathrm{E}[f_1(\boldsymbol{\pi}, \mathbf{v})|\pi_0]$, where $\boldsymbol{\pi}$ and $\mathbf{v}$ represent random variables whose values are distributed according to the exploration distribution $\mathcal{N}(\pi_0, \Sigma_e)$ and the output distribution respectively. The gradient with respect to $\pi_0$ is written as $\nabla_{\pi_0} \mathrm{E}[f_1(\boldsymbol{\pi}, \mathbf{v})|\pi_0] = A_\pi$ and therefore, a sensible gradient estimator returns an estimate of $A_\pi$ from the $n$ policy trials. The gradient estimator, which we denote by $g_1(\Pi, f)$ is given by the linear regression equation:

$$g_1(\Pi, f) = (\Pi^T \Pi)^{-1} \Pi^T f$$

The variance of the $g_1$ is written as

$$\begin{aligned} \mathrm{var}[g_1(\Pi, \mathbf{f})] &= (\Pi^T \Pi)^{-1} \Pi^T \mathrm{E}[\mathbf{v}\mathbf{v}^T] \Pi (\Pi^T \Pi)^{-1} \\ &= (\Pi^T \Pi)^{-1} \Pi^T ({A_s}^T \Sigma_s A_s + \sigma^2) \Pi (\Pi^T \Pi)^{-1} \\ &= (\Pi^T \Pi)^{-1} ({A_s}^T \Sigma_s A_s + \sigma^2), \end{aligned}$$

where $\mathbf{f}$ is a column vector random variable whose entries are distributed according to the output distribution. This quantity is for a fixed set of policies $\Pi$. The variance of the $g_1$ averaged over the randomness of the policies drawn for exploration purposes can be determined by observing that the distribution of the matrix $(\Pi^T \Pi)^{-1}$ is an inverted Wishart with $n$ degrees of freedom where $d$ is the number of policy parameters.

$$\mathrm{var}[g_1(\boldsymbol{\Pi}, \mathbf{f})] = \frac{{\Sigma_e}^{-1} ({A_s}^T \Sigma_s A_s + \sigma^2)}{(N - d - 1)}, \quad (3)$$

where $\boldsymbol{\Pi}$ is a random variable where each row is distributed according to the exploration distribution.

### 2.2.2 Include Sensor Data

A linear model that predicts the score as a function of both the policy parameters and sensor data will have the noise on the output partially explained by the sensor data. An estimate of the score as a linear function of the policy parameters and sensor data can be written as

$$\begin{bmatrix} g_2(\Pi, S, f) \\ \beta_2(\Pi, S, f) \end{bmatrix} = \begin{bmatrix} \Pi^T \Pi & \Pi^T S \\ S^T \Pi & S^T S \end{bmatrix}^{-1} \begin{bmatrix} \Pi^T f \\ S^T f \end{bmatrix},$$

where $\beta_2(\Pi, S, f)$ determines how the sensor values affect the perturbations in the score. The gradient of $\mathrm{E}[f_2(\boldsymbol{\pi}, \mathbf{s}, \mathbf{f})|\pi_0]$ with respect to $\pi_0$ is written as $\nabla_\pi \mathrm{E}[f_2(\boldsymbol{\pi}, \mathbf{s}, \mathbf{f})|\pi_0] = A_\pi$ and so we can use $g_2(\Pi, S, f)$, the first $d$ entries in the above vector, as our estimate of the gradient. The variance of the above expression is written as

$$\mathrm{var} \begin{bmatrix} g_2(\Pi, S, \mathbf{f}) \\ \beta_2(\Pi, S, \mathbf{f}) \end{bmatrix} = \begin{bmatrix} \Pi^T \Pi & \Pi^T S \\ S^T \Pi & S^T S \end{bmatrix}^{-1} \sigma^2.$$

We take the inverse of the Schur complement with respect to $S^T S$ to find the variance of $g_2$:

$$\mathrm{var}[g_2(\Pi, S, \mathbf{f})] = (\Pi^T \Pi - \Pi^T S (S^T S)^{-1} S^T \Pi)^{-1} \sigma^2.$$

This quantity is for a fixed set of policies $\Pi$ and sensor values $S$. The variance of $g_2$ averaged over different exploration policies and sensor values, assuming that the sensors are independent of both the policy parameters and output noise $w$, is given by the following equation:

$$\mathrm{var} \begin{bmatrix} g_2(\boldsymbol{\Pi}, \mathbf{S}, \mathbf{f}) \\ \beta_2(\boldsymbol{\Pi}, \mathbf{S}, \mathbf{f}) \end{bmatrix} = \begin{bmatrix} \Sigma_e & 0 \\ 0 & \Sigma_s \end{bmatrix}^{-1} \frac{\sigma^2}{(N - d - d_s - 1)}$$

$$\mathrm{var}[g_2(\boldsymbol{\Pi}, \mathbf{S}, \mathbf{f})] = \frac{{\Sigma_e}^{-1} \sigma^2}{(N - d - d_s - 1)}, \quad (4)$$

where $d_s$ is the dimensionality of the sensor data.

The expressions for the variance of the two gradient estimators (equations 3 and 4) differ from each other in two factors. The variance of the estimator that ignores the sensor data has a factor of $({A_s}^T \Sigma_s A_s + \sigma^2)$ which is reduced to $\sigma^2$ in the estimator that incorporates the sensor data. We see that we get bigger reductions whenever the sensor information provides more information about the score. The second difference between the two estimators favors the estimator that ignores the sensor data because the denominator in equation 3 has a term that is larger than the corresponding term in equation 4. The difference in the denominators is the dimensionality of the sensor data $d_s$, which suggests that we should choose sensor encodings of low dimensionality. Whether $g_2$ is more efficient than $g_1$ depends on the relative strength of these two factors.

#### 2.2.3 Correlated Sensors

If the sensors are correlated with the policy parameters then the gradient estimator that incorporates sensor data will be biased. In this situation we can represent the distribution over sensors as a linear Gaussian distribution $s \sim \mathcal{N}({A_{\pi,s}}^T \pi + b_{\pi,s}, \Sigma_s)$. Inserting this into score function $f_2$ gives the following equation:

$$f_3(\pi, s, w, w_{\pi,s}) = \pi^T A_\pi + \pi^T A_{\pi,s} A_s + {b_{\pi,s}}^T A_s + {w_{\pi,s}}^T A_s + b + w$$

where $w_{\pi,s}$ is a zero-mean Gaussian random variable with variance $\Sigma_s$. The gradient with respect to $\pi_0$ is written as $\nabla_{\pi_0} \mathrm{E}[f_3(\boldsymbol{\pi}, \mathbf{s}, \mathbf{w}, \boldsymbol{w_{\pi,s}}) | \pi_0] = A_\pi + A_{\pi,s} A_s$ which means that gradient estimator $g_2$ is biased by $A_{\pi,s} A_s$ whenever the sensors are correlated with the policy parameters. The variance of the estimator also changes in the case of correlated sensors:

$$\mathrm{var}[g_2(\mathbf{\Pi}, \mathbf{S}, \mathbf{f})] = \frac{(\Sigma_e - D)^{-1} \sigma^2}{(N - d - d_s - 1)}$$
$$D := \Sigma_{es} ({A_{\pi,s}}^T \Sigma_e A_{\pi,s} + \Sigma_s)^{-1} {\Sigma_{es}}^T,$$

where $\Sigma_{es} = \Sigma_e A_{\pi,s}$ is the covariance of the policy parameters and sensor values. Thus we see that it is best to choose sensor encodings where the sensor values are uncorrelated with the policy parameters.

### 2.3 FINDING GOOD SENSOR ENCODINGS

In the prior subsection we saw that $S$ should be independent of $\Pi$ to give an unbiased estimate of the gradient. This is often not true in many problems. The analyzes also suggests that to get an improved gradient estimator, we should prefer low-dimensional sensor encodings that have a great influence on the score. This section presents a heuristic that can be used to find good sensor encodings that give performance gains in problems where the assumptions do not hold.

We find good sensor encodings by taking sensor variables that lie in a high dimensional space and projecting them down to a low-dimensional subspace. Intuitively, we should prefer directions that maintain the influence of the sensor data on the score. We also want the sensor data to be uncorrelated with the policy parameters to minimize the bias. Our approach is to search over possible sensor encodings to find the optimal projection at each hill-climbing step. We use cross-validation to measure the quality of each projection. Let $\Pi_{\text{-}i}$ equal $\Pi$ with the $i$th row removed, let $S_{\text{-}i}$ equal $S$ with the $i$th row removed, and let $f_{\text{-}i}$ equal $f$ with the $i$th entry removed. Let $B$ be a matrix that projects the raw sensor data down to a low-dimensional subspace. We use a quasi-Newton method to minimize the following cost function:

$$J = \sum_{i=1}^{n} ({\pi_i}^T g(\Pi_{\text{-}i}, S_{\text{-}i} B, f_{\text{-}i}) + o(\Pi_{\text{-}i}, S_{\text{-}i} B, f_{\text{-}i}) - f_i)^2,$$

where $g$ estimates the gradient using $g_2$ after the inputs have been centered around zero and $o$ estimates the appropriate offset term. Thus we learn the gradient and corresponding offset terms using the data from a collection of policy trials with a single policy trial held out at a time. This gradient and offset is then used to predict the score of the held out sample. Given the optimal projection $B^*$ from the above procedure, we estimate the gradient using the full set of policy trials:

$$\nabla_\pi V(\pi)|_{\pi=\pi_0} \approx g_2(\Pi, SB^*, f). \qquad (5)$$

## 3 RESULTS

In this section we describe how an agent can use its sensor data to improve its learning performance on a dart throwing task [6] and a quadruped locomotion task. These partially observable sequential tasks are complicated partly due to the fact that the transition and sensor models and their structures are unknown. These properties are characteristic of a wide range of real-world tasks.

Figure 4(a) shows a single frame of the dart throwing problem where the objective is to throw a dart with minimal mean squared error (measured from where the dart hits the wall to the center of the dart board). The arm is modeled as a three-link rigid body with dimensions based on biological measurements [2]. The links correspond to the upper arm, forearm, and hand and are connected using a single degree of freedom rotational joint. The upper arm is connected to the shoulder at a fixed location. The arm is controlled by applying torques at each joint. These torques are generated by a PD-controller that attempts to move the arm through a desired trajectory, specified by a cubic spline for each joint angle. The starting posture of the arm is fixed in advance and the control policy specifies three spline knot positions for each joint, yielding 9 policy parameters in all. The controller is simulated for approximately 0.2 seconds and then the dart is released. The torques calculated by the PD-controller are perturbed by additive and multiplicative noise,[3] and there is Gaussian noise added to the release time with $\sigma = 0.01$.

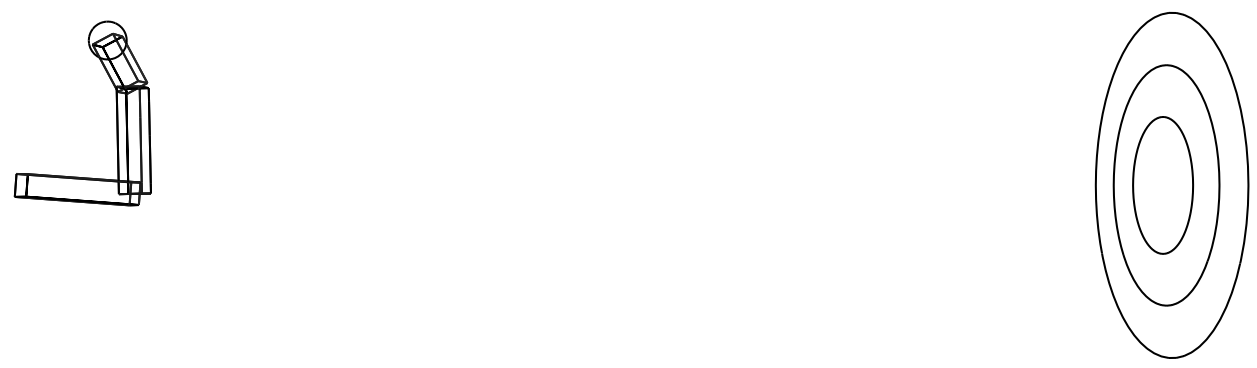

(a)

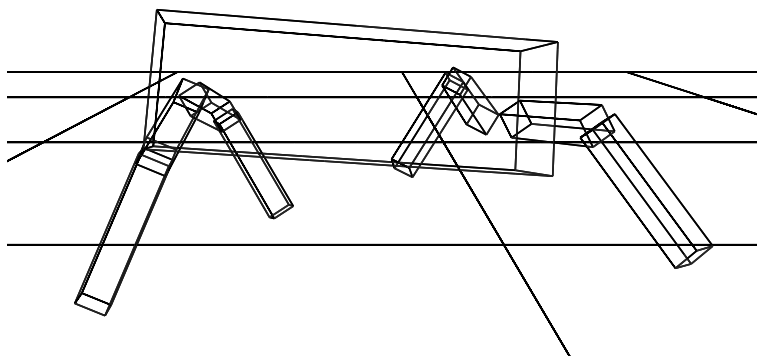

(b)

Figure 4: (a) Dart throwing problem (b) Quadruped locomotion problem.

Figure 4(b) shows a single frame of the quadruped locomotion problem where the objective is to maximize the sustained speed of the robot. Each leg of the quadruped has four degrees of freedom (three at the shoulder joints and one at the elbow). The quadruped is controlled by applying torques at each joint which gives the system 16 degrees of controllable freedom. The torques are generated by a PD-controller that attempts to move each leg through a desired trajectory, specified by a truncated Fourier Series for each joint angle. Each controllable degree of freedom has three corresponding parameters and the right side of the body is constrained to move the same as the left except offset by 180 degrees; there are 24 policy parameters. The controller is simulated for 3 seconds for each policy trial and the distance travelled is the objective function. As with the dart thrower, noise enters the system by perturbing the torques given by the PD-controller by additive and multiplicative noise.

[3]Multiplicative noise has been shown to explain some aspects of biological motion [4, 12].

### 3.1 SENSORS FOR MOTOR CONTROL

The systems described in the previous subsection are capable of measuring the state of the observable joint angles during each policy trial. In these two tasks, the observable joints are the same as the controllable joints. In the quadruped task, this means that while the agent can sense the positions of each leg relative to the body, it does not have access to the absolute position and rotation of the torso.

Our task is to take the sensed trajectories and to transform the values to something that gives us improvements in our gradient estimator. The sensor encodings should be independent of the policy parameters and so we attempt to find the difference between the observed motion of the system and the expected motion at each time step. The idea of using sensory data to cancel out the effect of one's own motion is also present in the biology literature [14]. We approximate this difference by learning a crude approximation to the dynamical system in a pre-processing phase. Using the joint-space representation of each system, the dynamics are governed by the following second-order nonlinear differential equation:

$$m(x)\ddot{x} = u(t) + g(x) + c(x, \dot{x}) + w(x, \dot{x}, t).$$

where $x$ is the physical state of the system, $m(x)$ is the joint-space inertia matrix, $u(t)$ are the forces and torques applied to the system, $g(x)$ is the gravitational force, $c(x, \dot{x})$ are the Centrifugal and Coriolis forces, and $w(t)$ is the noise plus any external forces (e.g., the ground pushing up on the feet of the quadruped). A discrete time version of this equation is written as:

$$m(x)a = u(t) + g(x) + c(x, v) + w(x, v, t).$$

where $v$ are the velocities and $a$ is the acceleration.

We approximate the expected acceleration by predicting the following quantities as a function of the observable states and velocities:

$$\begin{aligned} \text{vec}(m(x)^{-1}) &\approx A_M \phi(x_{(o)}) \\ \text{vec}(m(x)^{-1} g(x)) &\approx A_G \phi(x_{(o)}) \\ \text{vec}(m(x)^{-1} c(x, v)) &\approx A_C \phi(\left[x_{(o)}{}^T, v_{(o)}{}^T\right]^T), \end{aligned}$$

where $x_{(o)}$ and $v_{(o)}$ contain the observable components of the state and velocity terms and where $\phi(x) = \text{triu}([1, x^T]^T[1, x^T])$ is a function that augments its input with quadratic terms ($\text{triu}(X)$ is a function that returns the upper triangular part of $X$ stacked as a single column vector). We use linear regression to learn a model of each of these components as a function of the observable state variables. For example in the quadruped problem we will learn these linear models without regard to the absolute rotation of the system. This is clearly an approximation for the terms that involve gravity because the direction of the gravitational force, from a frame of reference attached to the torso, depends on its rotation relative to the ground frame.

We learn these parameters in a pre-processing stage by examining random states $(x_i, v_i)$ in the dynamical system and examining the mass matrix $m(x_i)$, the gravity forces $g(x_i)$, and the Coriolis and Centrifugal forces $c(x_i, v_i)$. The samples are drawn from a distribution of states that are likely to be encountered during policy execution. We learn the linear relationships using the following equations:

$$A_M := ({\Phi_1}^T \Phi_1)^{-1} \Phi_1^T M$$
$$A_G := ({\Phi_1}^T \Phi_1)^{-1} \Phi_1^T G$$
$$A_g := ({\Phi_2}^T \Phi_2)^{-1} \Phi_2^T C,$$

where

$$\begin{aligned}
\alpha_i &= \phi(x_{i(o)}) \\
\beta_i &= \phi([{x_{i(o)}}^T, {v_{i(o)}}^T]^T) \\
\Phi_1 &= [{\alpha_1}^T, \ldots, {\alpha_{n_f}}^T]^T \\
\Phi_2 &= [{\beta_1}^T, \ldots, {\beta_{n_f}}^T]^T \\
M &= [\mathrm{vec}(m(x_1)^{-1})^T, \ldots, \mathrm{vec}(m(x_{n_f})^{-1})^T]^T \\
G &= [\mathrm{vec}(m(x_1)^{-1} g(x_1))^T, \ldots, \\
&\quad \mathrm{vec}(m(x_{n_f})^{-1} g(x_{n_f}))^T]^T \\
C &= [\mathrm{vec}(m(x_1)^{-1} c(x_1, v_1))^T, \ldots, \\
&\quad \mathrm{vec}(m(x_{n_f})^{-1} c(x_{n_f}, v_{n_f}))^T]^T.
\end{aligned}$$

During each policy execution we can take these parameters to estimate the expected acceleration at each time step as follows:

$$\begin{aligned}
\hat{a}(\pi, t, x_{(o)}, v_{(o)}) = \; & \mathrm{resh}_M(A_M \phi(x_{(o)})) u(\pi, t) + \\
& \mathrm{resh}_G(A_G \phi(x_{(o)})) + \\
& \mathrm{resh}_C(A_C \phi([{x_{(o)}}^T, {v_{(o)}}^T]^T)),
\end{aligned}$$

where the resh function reshapes a matrix to its original size. The difference in velocity is computed as the actual velocity at each time step $v_{(o)}(t)$ minus the velocity predicted using the following expression:

$$v_{(o)}(t) \approx v_{(o)}(t-1) + \hat{a}(\pi, t, x_{(o)}(t-1), v_{(o)}(t-1)) \Delta t,$$

where $\Delta t$ is the time between sensor measurements.

We reduce the number of sensor values by projecting these difference curves down onto a set of basis functions. We chose the same basis functions that we used to encode the policy for the two tasks; we used splines for the dart and a truncated Fourier series for the quadruped. This is the sensor data that we give to the gradient estimator algorithms as described in Section 2.2. We also include a sensor that detects the release time for the dart throwing task.

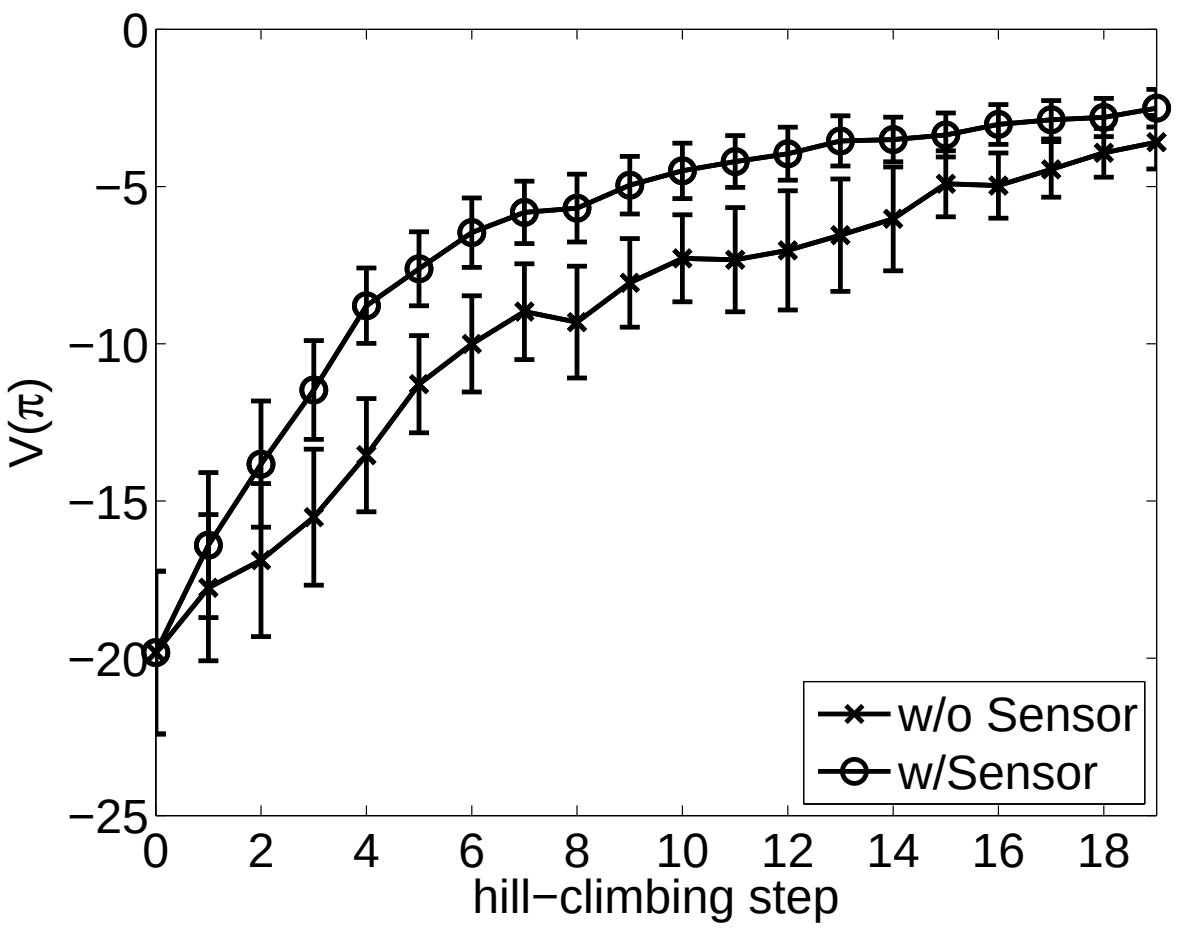

Figure 5: The learning curve performance of two policy search strategies in the dart throwing domain. At each hill-climbing step we drew a single sample from 12 different policies and we averaged over 48 hill-climbing runs.

### 3.2 LEARNING PERFORMANCE

Figure 5 shows the learning curves for the dart throwing problem. We get a substantial improvement in the learning performance when using an algorithm that incorporates sensor data when compared to an algorithm that ignores the sensor data. At each hill-climbing step we drew a single sample from 12 different policies and we averaged over 48 hill-climbing runs. The policies for each hill-climbing step were drawn from a Gaussian distribution for exploration purposes.

Figure 6 shows the learning curves for the quadruped locomotion problem. We get an improvement in the learning performance when using of an algorithm that incorporates sensor data when compared to an algorithm that ignores the sensor data. At each hill-climbing step we drew a single sample from 30 different policies and we averaged over 48 hill-climbing runs. The policies for each hill-climbing step were drawn from a Gaussian distribution for exploration purposes.

## 4 DISCUSSION

In this paper we demonstrated how one may incorporate sensor data into the gradient estimation task to improve the performance of policy search. We showed that the level of improvement depends on the amount of information that the sensors provide about the noise-induced perturbations on the score. We also showed that the performance gains depend on the dimensionality of the sensor data. It is important to choose sensor encodings that are independent of the policy parameters to minimize the bias. We also presented a technique to find good sensor encodings for problems in which these assumptions (low-dimensionality and statistical indepedence) do not hold. Finally, we presented learning curves that show improvements in the learning performance for a toy cannon problem, dart throwing task, and quadruped locomotion task.

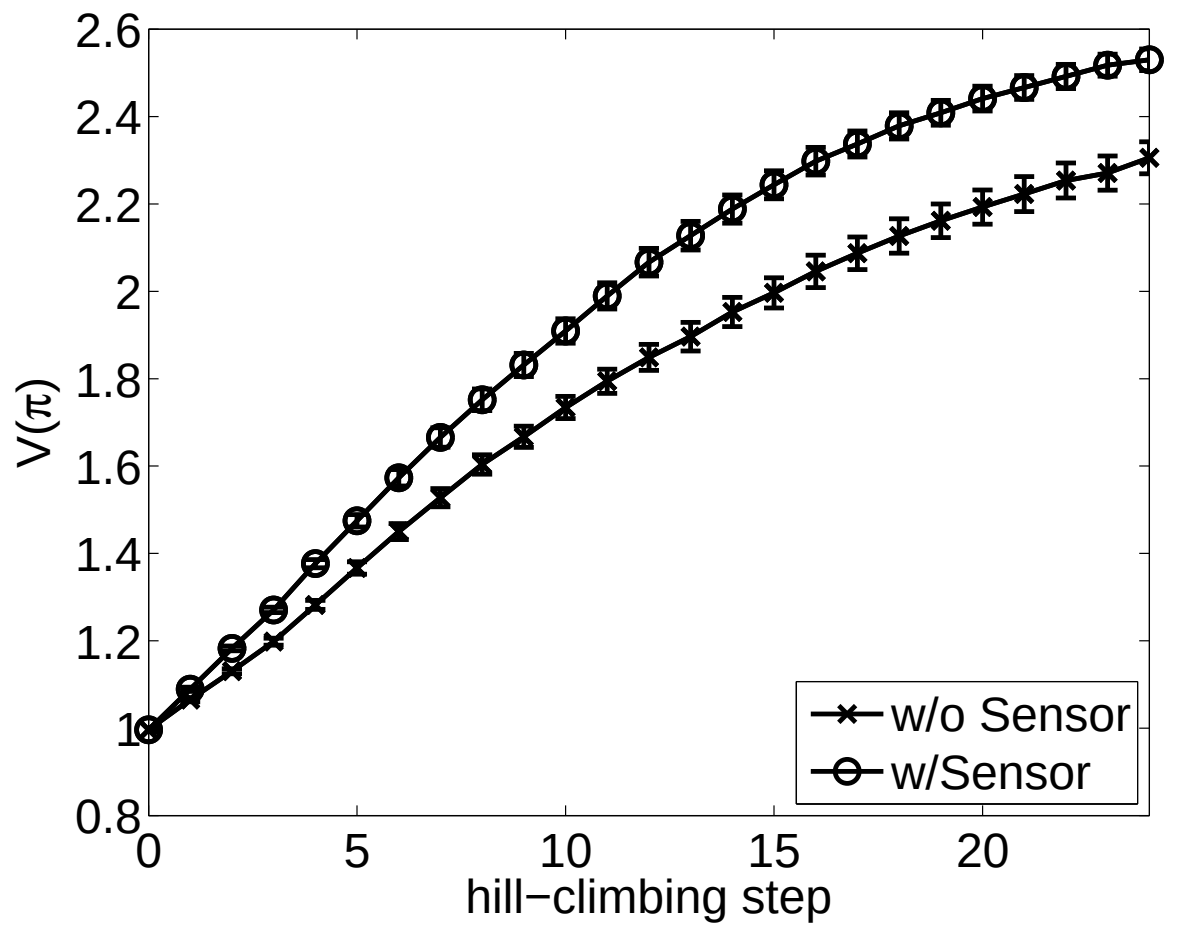

Figure 6: The learning curve performance of two policy search strategies in the quadruped learning domain. At each hill-climbing step we drew a single sample from 30 different policies and we averaged over 48 hill-climbing runs.

In this paper the distribution of every random variable was approximated by using a linear Gaussian relationship. Improvements in performance may be realized if we use non-linear mappings. Other improvements may come from incorporating our knowledge of the physics behind each task. For example, in the cannon problem we already know the equations of projectile motion. Thus, given the actual controls, we should be able to accurately predict the score. Even in cases in which we do not know the equations of motion, we often know qualitative information about the motion, such as the fact that increasing the desired velocity causes the cannon ball to fly further (i.e., the distance travelled is monotonically increasing as a function of the desired velocity). One possible approach to incorporating this information is to place a prior on the parameters that reflects these constraints.